\documentclass[11pt]{article}

\usepackage[final]{acl}
\usepackage{longtable}
\usepackage{latexsym}

\usepackage[english,bidi=default]{babel}
\babelfont{rm}[
  BoldFont=texgyretermes-bold.otf,
  ItalicFont=texgyretermes-italic.otf,
  BoldItalicFont=texgyretermes-bolditalic.otf
]{texgyretermes-regular.otf}

\babelprovide[import]{hebrew}
\babelfont[hebrew]{rm}[
  BoldFont=DavidCLM-Bold.otf
]{DavidCLM-Medium.otf}

\usepackage{microtype}

\usepackage{graphicx}
\usepackage{booktabs}
\usepackage{enumitem}

\usepackage{stfloats}

\title{JudgeMeNot: Personalizing Large Language Models to Emulate Judicial Reasoning in Hebrew}
\author{Itay Razumenko, Arnon Sturm, \and Nir Grinberg \\
  Computer and Information Science \\
  Ben-Gurion University of the Negev \\
  Beer-Sheva, Israel \\
  \texttt{itayraz@post.bgu.ac.il, \{sturm, nirgrn\}@bgu.ac.il}}

\begin{document}
\maketitle
\begin{abstract}
Despite significant advances in large language models, personalizing them for individual decision-makers remains an open problem. 
Here, we introduce a synthetic-organic supervision pipeline that transforms raw judicial decisions into instruction-tuning data, enabling parameter-efficient fine-tuning of personalized models for individual judges in low-resource settings. 
We compare our approach to state-of-the-art personalization techniques across three different tasks and settings. 
The results show that Causal Language Modeling followed by synthetically generated instruction-tuning significantly outperforms all other baselines, providing significant improvements across lexical, stylistic, and semantic similarity.
Notably, our model-generated outputs are indistinguishable from the reasoning of human judges, highlighting the viability of efficient personalization, even in low-resource settings.
\end{abstract}

\section{Introduction}
Human writing is characterized by stable individual style and reasoning, which shape the way arguments are framed and supported~\cite{koppel2009computational}. 
In the legal domain, this is particularly acute: judicial decisions are not merely a mechanical application of the law to a case, they reflect judge-specific patterns in reasoning, emphasis, and rhetorical structure~\cite{ash2024relatio,sunstein2006judges}. 
Nevertheless, most Legal NLP work, from outcome prediction to summarization, treats judges as interchangeable decision-makers. 
This approach reduces profound interpretive differences to mere noise and effectively misses the opportunity to build a deeper understanding of the variation in judicial reasoning and style.

\begin{figure}[t]
  \centering
  \includegraphics[width=1\linewidth]{figures/fig1.pdf}
  \caption{Illustrative example: Two personalized LLMs trained (top) to emulate two human judges, and then queried (bottom) with a question inspired by Muscarello v. United States. The query prompts different philosophies for interpreting the law: a textualist approach, which focuses on the literal meaning, and a purposivist approach, which focuses on legislative intent.}
    \label{fig:judge_example}
\end{figure}

Despite significant advances in reasoning models \citep[e.g.,][]{guo2025deepseek}, research on personalizing for individual decision-makers remains nascent.
Recent surveys highlight growing interest in personalization for content generation~\cite{zhang2024_personalization_survey,xu-etal-2025-personalized}, yet existing efforts focus on user preferences (e.g., style, product recommendations) rather than reasoning. 
A major barrier is the need for objective verification signals (e.g., code or math proofs) to reward correct reasoning steps~\cite{lightman2024let}. 
While traces of individual human reasoning are generally sparse, the legal domain provides an abundance of such data, making it an ideal testbed for personalizing large language models that enable reasoning. 
Judges tackle tough decisions on a regular basis~\cite{posner2008judges,epstein2013behavior}, and write long and detailed texts justifying them. 
By decomposing verdicts into their constituent parts, rather than focusing only on the final ruling, one can extract many reasoning decisions and thousands of these recorded decisions throughout a judge's tenure on the bench.
Furthermore, tackling personalization in a low-resource language provides a rigorous stress test for transferability beyond standard English benchmarks~\cite{guha2023legalbench}.

Bridging these gaps is critical for addressing the open challenge of ``deliberative reasoning'' in personalization, where current methods struggle to model complex, slow-thinking cognitive processes~\cite{xu-etal-2025-personalized}.
Personalization offers a path to combat linguistic homogenization and ``mode collapse'' of standard pretraining~\cite{guo2025benchmarking}.
In the legal domain, this is not merely a stylistic preference but a functional necessity. 
Judicial decision-making is inherently interpretive, requiring judges to balance competing values (e.g., deterrence vs. rehabilitation) and philosophies~\citep[][see example in Figure~\ref{fig:judge_example}]{posner2008judges, sunstein2006judges}.
Treating judges as interchangeable ignores the primary driver of legal outcomes.
Therefore, personalization is crucial for high-fidelity counterfactual analysis and trust by legal scholars seeking to examine a fuller range of possible interpretations~\cite{zhang2024_personalization_survey}. 

However, several challenges make it particularly hard to train high-fidelity LLMs for individual judges. 
First, the raw data is not structured for reasoning supervision.
Verdicts are lengthy, unstructured documents where reasoning is often inextricably woven with procedural boilerplate, recitation of the facts, and personal circumstances~\cite{vsavelka2018segmenting}. 
Second, even when identified, judges' decisions often appear ``unprompted'' in the text, without the implicit internal question that triggered them. 
Third, it is unclear whether the reasoning traces from a single judge will be strong and consistent enough to move the priors of the base model, while keeping model training computationally efficient enough to enable scaling up to hundreds or even thousands of individuals. 

In this study, we introduce a scalable framework for personalizing LLMs for judicial reasoning and style without manual annotation (see Figure~\ref{fig:judge_example} for an illustrative example). 
To overcome the scarcity of supervision, we propose a synthetic-organic alignment pipeline that transforms raw judicial decisions into instruction-tuning data. 
Using an agentic workflow, which we manually validated, our approach identifies reasoning statements in verdicts and generates synthetic questions that plausibly prompted them. 
This allowed us to train personalized models through parameter-efficient fine-tuning (PEFT) on a reasoning-focused instruction set and compare against other personalization techniques, namely standard Causal Language Modeling (CLM) and Retrieval-Augmented Generation (RAG), across configurations and model combinations. 
We test our models on three different tasks -- next token prediction, question-answering, and author discernment -- and evaluate the lexical, stylistic, and semantic fidelity of leading LLM personalization techniques.

\noindent Therefore, our contributions are the following:
\begin{itemize}[nosep, leftmargin=*]
\item A benchmark for evaluating personalized models in text generation and reasoning tasks in the morphologically rich, low-resource language of Hebrew.
\item A synthetic-organic alignment pipeline that generates reasoning-focused instruction pairs.
\item A parameter-efficient methodology for training personalized LLMs to emulate the reasoning and style of individual judges. 
\item A systematic evaluation of LLM personalization techniques across multiple tasks, including an ablation study of the dominant factors for improvement. 
\end{itemize}

\section{Related Work}

The current study builds on three lines of work:  personalizing large language models, reasoning models in various tasks, and synthetic question-generation for enhanced model supervision. 

\subsection{Personalization of LLMs}
\noindent\textbf{Parameter-Efficient Fine-Tuning (PEFT)} is a central approach to personalization that adapts an LLM by training only a small subset of parameters or by inserting lightweight trainable modules \cite{houlsby2019parameter,hu2022lora}. OnePeFTPerUser combines PEFT with retrieval capabilities to enable generation based on user-specific evidence, resulting in gains on labeled personalization tasks such as tagging and classification \cite{tan2025_oppu_ppeft}. \citet{liu-etal-2024-customizing-large} fine-tune LLaMA-2 to improve lexical and syntactic alignment with the specific authors. Comparative studies across domains further report that LoRA and QLoRA often offer a strong tradeoff between parameter efficiency and adaptation quality, while prompt tuning is typically lighter but weaker on more complex tasks \cite{gajulamandyam2025domain}. However, applying PEFT to long, unstructured legal texts under limited supervision, and evaluating whether it captures judge-specific \emph{reasoning} (not only style), remains underexplored \cite{zhang2024_personalization_survey}.

\noindent\textbf{Retrieval-Augmented Generation (RAG)} supports personalization by retrieving user- or domain-specific material at inference time. Style-aware variants aim to capture recurring authorial patterns beyond lexical overlap \cite{neelakanteswara-etal-2024-rags}, and in law, CBR-RAG combines case-based reasoning with retrieval to improve factual grounding in legal question answering \cite{wiratunga2024_cbr_rag_legal}. However, these approaches focus on accurate answers rather than the recurring reasoning of a specific judge.

\noindent\textbf{Prompt-based personalization.}
Prompting can encode user or domain cues directly in the input and improve alignment with preferences \cite{oba2023_perplm}, but profile-based prompts are often noisy, and prompting alone may not yield stable user-specific outputs. \citet{wang-etal-2025-catch} and  \citet{gajulamandyam2025domain} show that LLMs still struggle to imitate implicit writing styles, and prompt tuning often underperforms stronger adaptation methods in complex domain-specific settings. Overall, personalization to a \emph{specific individual}, such as a particular judge, remains underexplored, especially under limited supervision.

\noindent\textbf{LLM Simulation.} A growing body of research leverages LLM personalization methods to emulate the language, preference, and behavior of individuals or personas across different contexts~\cite{shao-etal-2023-character,li2023chatharuhi,park-etal-2025-charactergpt,dinu-etal-2025-analyzing,park2024generative}. 
Our work complements this line of work by focusing on reasoning as a process, extracting and modeling granular decisions throughout a case, and not the final ruling. 

\subsection{Reasoning Models in Verifiable Domains}
Reasoning-oriented post-training has recently improved LLMs on multi-step problems, particularly when solutions are externally checkable (e.g., math, code, and scientific reasoning) \citep{guo2025deepseek}. Prior work distinguishes several reasoning modes, including deductive, inductive, and abductive reasoning, as well as analogical and causal reasoning \citep{huang-chang-2023-towards}. When the correctness of steps is well-defined, process supervision can further reduce logical errors by supervising intermediate steps \citep{lightman2023let}. Judicial reasoning does not offer the same objective supervision.
In this work, we evaluate LLM emulation of judges' rationale, rather than step correctness, and use this signal as evidence that judge-specific reasoning patterns are learnable.

\subsection{Synthetic Question-Generation}
To enable personalization of reasoning models, one needs supervision signals. Synthetic supervision is a common strategy for expanding training signals when manual annotations are not feasible. In this work, we  generate question-answer pairs to emulate reasoning activities. Prior work on question-answer generation shows that generated QA pairs can support adaptation and provide structured evaluation material by encouraging close reading and explicit answers \cite{ushio-etal-2023-empirical,chen-etal-2024-reinstruct,nayak-etal-2024-learning}. GPT-based methods were shown to be effective in annotating and classifying legal texts with little or no task-specific training. For example, \citet{savelka2023unlocking, savelka2023can} demonstrates that GPT-4 and GPT-3.5 can match trained human annotators on a sentence-level legal analysis task when guided by detailed annotation instructions. 

Nevertheless, there remains a gap in extracting judicial justifications and turning them into clear question–answer pairs tailored to a specific decision-maker. Our work moves toward filling this gap by extracting judge-specific reasoning and generating synthetic questions for it, enabling judge-level modeling even when data are limited.

\section{Methodology}
In this section, we describe how the data were curated and generated, the personalization methods we experimented with, and the evaluation metrics used to assess their success.

\subsection{Data Curation and Generation}
\label{sec:data}

\textbf{Judicial Decisions.}
We collected from the Nevo Legal Database\footnote{\url{https://www.nevo.co.il}} public domain verdicts by the magistrate and district courts in Israel. We filtered for single-judge rulings to ensure unambiguous authorship, and used only verdicts where we could reliably identify the judge's summary judgment section and exclude other sections (recitation of facts, external circumstances, etc.). 
To ensure models have sufficient training data, we focused on judges with at least 100 such summary judgments. 
The final dataset comprises summary judgments of 29 judges, including their written opinions spanning multiple years.

\noindent\textbf{Question Generation.} To create high-quality synthetic-organic reasoning pairs, we construct an agentic workflow using multiple LLM agents for extraction and validation, applied over several iterations. In particular, we use GPT-4.1-mini for reasoning extraction (temperature 0.3) and GPT-4o-mini for validation (temperature 0.1), which we empirically found to perform best. Custom prompts were engineered to extract reasoning statements, validate the quality of this extraction, generate questions, and validate their fidelity. The entire pipeline, including all prompts, is described in Section~\ref{sec:qa-gen}. The overall process resulted in a set of 62{,}051 reasoning sentences and synthetic questions generated for them. 

We validated the quality of output generated by the aforementioned agentic workflow by having two annotators manually label a random sample of 175 instruction pairs. They independently assessed whether the organic reasoning statement extracted contained explicit judicial reasoning and whether the generated question was sufficiently related and could plausibly induce the reasoning statement. Inter-coder reliability was relatively high, with Gwet's AC1 of 0.75 for the statement extraction and 0.90 for question generation. In the randomly-sampled sample, 83\% of answers contained explicit reasoning, and 91\% of questions were semantically aligned with them. Both the raw verdicts and the reasoning instruction sets are publicly available\footnote{\url{https://doi.org/10.7910/DVN/8CYHIC}}.

\subsection{Personalization Methods}
Next, we evaluate leading approaches for LLM personalization, as well as, chaining of different models, as detailed below. 
All of our experiments are using open-source foundation models and publicly available code\footnote{\url{https://github.com/Socially-Embedded-Lab/JudgeMeNot}}.

\noindent\textbf{CausalLM Fine-Tuning with QLoRA:}
We fine-tuned multilingual Gemma~3 models~\cite{team2025gemma}, both 1B and 4B variants, using QLoRA~\cite{dettmers2023qloraefficientfinetuningquantized}.
We choose Gemma~3 because it is open, multilingual, and efficient for training on a per-judge basis in constrained computing resources environments. See Appendix~\ref{sec:appendix_training} for precise information about the computing resources used and training hyperparameters.
Causal language modeling (CLM) was applied to all raw, unlabeled summary judgment text from a single judge, separately training a LoRA adapter for each judge in the dataset while keeping the base Gemma weights frozen.
This training procedure encouraged the model to learn judge-specific writing patterns without any manual labeling.

\noindent\textbf{Instruction Fine-Tuning with QLoRA:}
We perform a per-judge instruction fine-tuning with QLoRA using the synthetic-organic reasoning pairs described in the previous section.
This training aims to focus model weights on reasoning statements.
Again, only the adapters were kept for each judge. 

\noindent\textbf{Chain-of-LoRA (CLM → Instruction Tuning):} 
Following the ideas of~\citet{xia2024chainloraefficientfinetuning}, we trained a Chain of LoRA (CoLA) model, where adapter weights are merged back into the model, updating the ``starting point'' before training each new adapter layer.
Specifically, we trained the CLM adapter, merged this into the Gemma base weights, then run a second QLoRA fine-tuning on the instruction set.
The goal of this procedure is to first adapt the model to the judge's general writing style, then specialize it with supervised reasoning examples.
We provide the full training configurations in Appendix~\ref{sec:appendix_training}.

\noindent\textbf{Retrieval-Based Personalization:}
Finally, we evaluated retrieval-based personalization that involves no training.
For each judge, we index the judge-specific instruction pairs, 
retrieve at inference time the top-$k$ most similar pairs for an input question, and include them in the prompt as in-context examples for generation. 
We experimented with generation using Gemma-3 and Gemini-3-Pro, which currently is one of the strongest reasoning models across leaderboards~\cite{gemini3pro_modelcard}. 

\subsection{Evaluation Metrics}
We used common metrics across two of our main tasks: next-token prediction and question-answering. 
Specifically, we used BLEU and ROUGE-1/2/L for lexical similarity, BERTScore~\cite{zhang2019bertscore} for semantic similarity, and Jensen--Shannon divergence over Part-of-Speech (POS) tags for stylistic similarity. 
Following common practice in author discernment~\cite{casal2023can,moro2026legal,zhang2024_personalization_survey}, we use classifier accuracy on held-out set, as detailed in \autoref{sec:exp3_authorship}). 
We use the latest Hebrew encoder architecture of DictaBERT~\cite{shmidman2023dictabert} as the underlying model for indexing documents for retrieval, computing BERTScore, performing Part-of-Speech (POS) tagging, and training our authorship classifier.

\section{Experimental Design}

We evaluate our personalized models on three tasks: (i) next-token prediction, (ii) question-answering, and (iii) author discernment. 
For next-token prediction, we used the first 15\% of a summary judgment as a seed for generating the next tokens.
For question-answering, we used the synthetic-organic instruction pairs. 
For author discernment, we trained a dedicated classifier per judge, as detailed below. 

\subsection{Next-token Prediction and Question Answering}

For both tasks, we follow the same procedure. For each judge $j \in \{1,\dots,J\}$, we construct a held-out test set $\mathcal{T}_j$.
We evaluate non-personalized baseline models on $\mathcal{T}_j$ and compare them to the personalized model $M_j$, trained only on judge $j$. 
All models were evaluated with the same prompts and decoding settings.
This provides a within-judge comparison that tests whether personalization improves generation quality on the judge's own data.

\textbf{Models:} We compare the Chain-of-LoRA model against several non-personalized baselines evaluated on every judge test set.
We include the base Gemma model (the foundation used in our personalization pipeline) as a direct reference point without judge-specific adaptation. We also evaluate DictaLM3 as a strong open Hebrew LLM baseline \cite{Shmidman2025DictaLM3}.
In addition, we report two multi-judge training baselines: \textbf{Joint-Gemma}, trained once on the union of all judges' training data, and \textbf{Mixed-Gemma}, trained on a size-matched mixture across judges where the total training budget is comparable to the average single-judge dataset.
We also compared the CoLA model with respect to other personalization models we trained (CLM and Instruction-tuning). 

To further check whether personalization is judge-specific rather than a generic improvement, we evaluate every personalized model on every judge's held-out test set.
For each judge $j$, we compare the performance of the model trained on judge $j$ (denoted $M_j$) on judge $j$'s test set $\mathcal{T}_j$ against the performance of models trained on other judges on the same test set.

Formally, for each metric we compute the test-set centered gap $\Delta_j = r_{jj} - \frac{1}{J-1}\sum_{k\neq j} r_{kj}$.

where $r_{kj}$ is the score of model $M_k$ evaluated on test set $\mathcal{T}_j$.
For metrics (such as ) higher scores of the judge's model is considered better, $\Delta_j > 0$ indicates that the judge-specific model performs better on the judge $j$'s own test set than models personalized to other judges, consistent with judge-specific adaptation.
We report the distribution of $\Delta_j$ across judges.

\subsection{Authorship Discernment}
\label{sec:exp3_authorship}
We assess judge-specific stylistic fidelity with a judge-specific binary classifier.
For each judge $j$, positive examples are real judge-authored reasoning sentences, and negative examples are either other-judge sentences (real-vs-real) or generated sentences from the judge's trained models.
We keep the classifier fixed across settings and report accuracy, with the random guessing and real-vs-real setting serving as reference points.

\begin{table}[t]
\centering
\small
\setlength{\tabcolsep}{4pt}
\resizebox{\columnwidth}{!}{%
\begin{tabular}{lcccc}
\toprule
Model & BLEU $\uparrow$ & BS-F $\uparrow$ & R-L $\uparrow$ & POS $\downarrow$ \\
\midrule
\multicolumn{5}{l}{\textbf{Non-personalized baselines}} \\
\hspace{3mm}Vanilla-DictaLM3 & 11.845 & 0.089 & 0.188 & -0.005 \\
\hspace{3mm}Vanilla-Gemma    & 11.997 & 0.113 & 0.192 & -0.017 \\
\hspace{3mm}Mixed-Gemma      & 9.205  & 0.043 & 0.083 & -0.012 \\
\hspace{3mm}Joint-Gemma      & 8.912  & 0.037 & 0.078 & -0.009 \\
\addlinespace
\multicolumn{5}{l}{\textbf{Retrieval-augmented baselines}} \\
\hspace{3mm}Gemini-3-Pro & 8.773 & 0.021 & 0.077 & -0.001$^{*}$ \\
\hspace{3mm}Gemma-3     & 10.421 & 0.023 & 0.148 & +0.011 \\
\addlinespace
\multicolumn{5}{l}{\textbf{Alternative personalization models}} \\
\hspace{3mm}Pers-CLM         & 11.748 & 0.079 & 0.181 & -0.001$^{*}$ \\
\hspace{3mm}Pers-IT          & 4.973  & 0.020 & 0.043 & -0.001$^{*}$ \\
\bottomrule
\end{tabular}%
}
\caption{The differences among the CoLA and the competitors (mean across judges) in the question-answer completion task. The CoLA outperforms the baseline, and the differences were statistically significant ($p<0.05$) \textbf{except} for the entries marked with an (*).}
\label{tab:qa_vs_all_models_compact}
\end{table}

\section{Results}
In this section, we present the question-answering results, whereas the analysis of the next-token prediction task is presented in Tables \ref{tab:clm_vs_all_models_appendix} and \ref{tab:cross_judge_all_methods}. 
The results reported here refer to the 4B models, as they consistently outperform their 1B counterparts, as appear in Table \ref{tab:4b_vs_1b_wilcoxon}.

\subsection{Question Answering}

Table \ref{tab:qa_vs_all_models_compact} summarizes the results of evaluating the improvements of personalization when referring to question generation. The Chain-of-LoRA model yields consistent and statistically significant gains over all non-personalized baselines on overlap- and semantic-oriented metrics (BLEU, ROUGE-L, and BERTScore-F).
The improvements are especially pronounced relative to the vanilla baselines, indicating that judge-level adaptation provides benefits beyond generic legal generation.
Across alternative personalization paradigms, the model attains the strongest overall performance, with clear gains over CLM-only and RAG-only personalization and a smaller but consistent advantage over instruction-tuning-only personalization.
For JSD-POS (lower is better), the Chain-of-LoRA model reduces divergence relative to most baselines, but not relative to the RAG personalization variant, where JSD-POS increases.
This suggests that improvements in semantic and lexical fidelity do not always translate to closer part-of-speech distributions, and that retrieval-based personalization can affect syntactic regularities differently than parameter adaptation.

On the Next Token Prediction evaluation, the CLM-based personalization variant performs best overall, outperforming the other training strategies on most metrics. We report the full CLM comparison in Table~\ref{tab:clm_vs_all_models_appendix} in the Appendix.

\begin{figure}[t]
  \centering
  \includegraphics[width=\columnwidth]{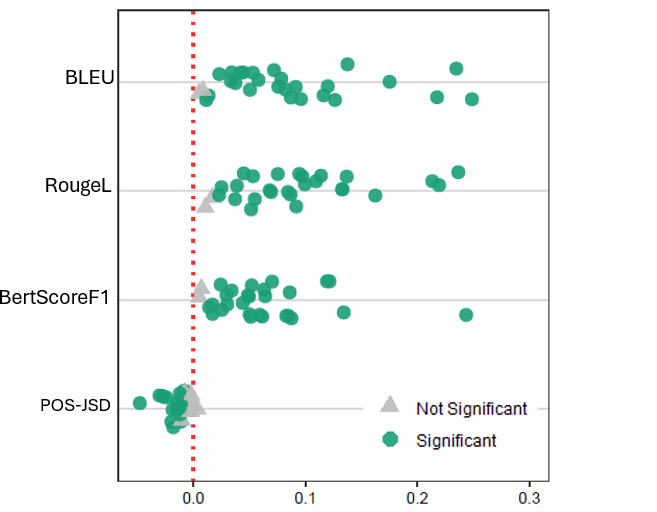}
\caption{
Each point shows the performance gap between the matched \textbf{CoLA-4B} model for a judge and CoLA-4B models trained on other judges, evaluated on that judge’s test set. Points to the \textbf{right} of zero indicate better performance for the matched model (for \textbf{POS-JSD}, where lower is better, the direction is reversed). \textbf{Green circles} indicate significant gaps ($p<0.05$); \textbf{grey triangles} are non-significant. \textit{BLEU is scaled ($/100$) for visualization.}
}

  \label{fig:forest-onecol}
\end{figure}

Figure~\ref{fig:forest-onecol} summarizes cross-judge specificity of  \textbf{Chain-of-LoRA (4B)} model using the test-set centered gaps $\Delta_j$.
For lexical and semantic metrics (BLEU, BERTScore-F, and ROUGE), $\Delta_j$ is predominantly positive, indicating that the model personalized to judge $j$ tends to outperform models trained on other judges when evaluated on $\mathcal{T}_j$.
This pattern suggests that the gains observed in the baseline comparison are not merely generic improvements: they are strongest when the model and test set correspond to the same judge.

For the structural metric JSD-POS, the differences are closer to zero and more variable across judges.
This indicates that personalization more consistently improves content fidelity and lexical similarity than it matches part-of-speech distributions, which may be shaped by shared institutional phrasing and boilerplate.
Full cross-judge results for all methods appear in  Table~\ref{tab:cross_judge_all_methods} in the Appendix.

\begin{table}[t]
\centering
\small
\setlength{\tabcolsep}{5pt}
\resizebox{0.8\columnwidth}{!}{%
\begin{tabular}{lc}
\toprule
Setting / Method & Acc. (\%) \\
\midrule
Random classifier          & 50.0 \\
Ground truth (real vs other judges) & 84.3$^{*}$ \\
\addlinespace

\multicolumn{2}{l}{\textbf{Non-personalized baselines}} \\
\hspace{3mm}Vanilla-Gemma-3       & 70.3$^{*}$ \\
\hspace{3mm}Vanilla-DictaLM-3     & 71.9$^{*}$ \\
\hspace{3mm}Mixed-Gemma-3         & 76.2$^{*}$ \\
\hspace{3mm}Joint-Gemma-3         & 77.0$^{*}$ \\
\addlinespace

\multicolumn{2}{l}{\textbf{Retrieval-augmented baselines}} \\
\hspace{3mm}Gemma-3               & 70.4$^{*}$ \\
\hspace{3mm}Gemini-3-Pro          & 73.8$^{*}$ \\
\addlinespace

\multicolumn{2}{l}{\textbf{Personalized models}} \\
\hspace{3mm}CLM                   & 56.2$^{*}$ \\
\hspace{3mm}IT                    & \textbf{49.6} \\
\hspace{3mm}CoLA                  & \textbf{49.8} \\
\bottomrule
\end{tabular}
}
\caption{
Authorship discernibility task: Classifier accuracy. across judge-specific classifiers.
An (*) marks significant differences from chance (50\%, $p<0.05$, paired Wilcoxon test across judges).
}
\label{tab:authorship_results_full}
\end{table}

\subsection{Authorship Discernment}

To assess judge-specific fidelity, we train a binary authorship classifier for each judge, distinguishing that judge's real reasoning sentences from a negative set.
We keep the positive class fixed (real sentences by judge $j$) and vary the negative class across settings.
Table~\ref{tab:authorship_results_full} reports mean accuracy across 29 judges and compares each setting to a random classifier.
In the ground-truth setting, where negatives are real sentences written by other judges, the classifiers achieve high average accuracy (0.843). This confirms that judges are separable in authentic text.  
When generating answers using the non-personalized or retrieval-augmented baselines, the classifier was able to detect differences between the sentences originating from the judges to their generated content (accuracy ranges between 70\% and 76.2\%). This implies that the baseline approaches failed to mimic the judges' fidelity. 

When generating answers using the personalized methods, results differ sharply by method.
CLM-only generations remain detectable: accuracy rises to 56.2\%, significantly above random, indicating that the classifier can still distinguish generated text from the judge's authentic writing.
In contrast, both instruction-tuning-only and Chain-of-LoRA reduce accuracy to random level (0.496 and 0.498, respectively; $p{>}0.8$), suggesting that these generations are difficult to distinguish from real judge sentences under this authorship test.
Taken together, these findings complement the semantic and cross-judge results by showing that the strongest personalization methods also suppress detectable authorship cues.

We observe a similar pattern when running the same task under the next-token prediction (CLM) training setting: model accuracies deviate from the random baseline, yet remain far from the ground-truth separability achieved with real other-judge negatives. Full results are reported in Table~\ref{tab:acc_vs_random_appendix} in the Appendix.

\subsection{Ablation Study}

\begin{figure}[t]
    \centering    \includegraphics[width=\columnwidth]{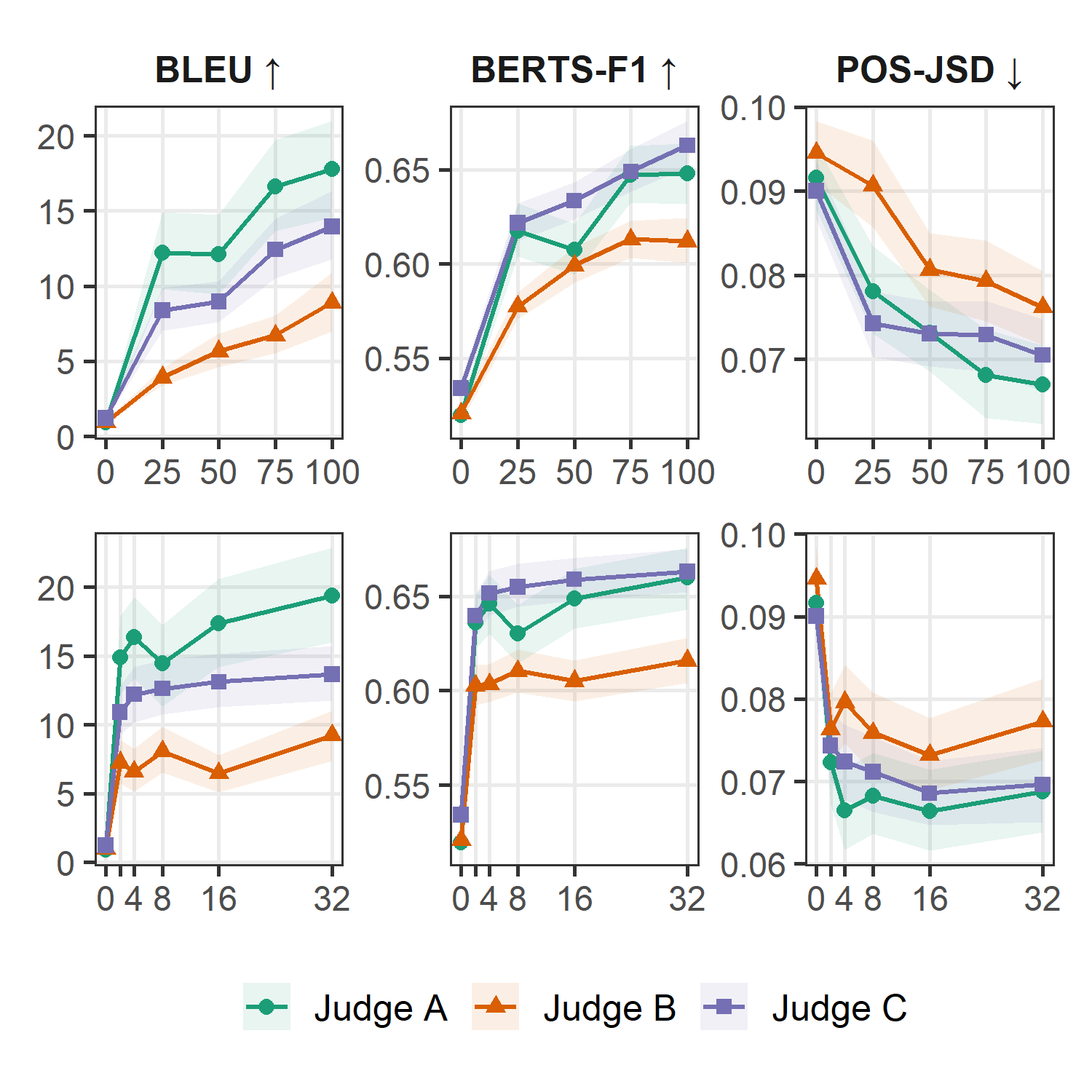} 
     \caption{
    Ablation study on three judges.
    Top row: performance as a function of training set size (x-axis: 25, 50, 75, 100\% of the judge’s training data).
    Bottom row: performance as a function of LoRA rank (x-axis: $r\in\{2,4,8,16,32\}$).
    }
    \label{fig:ablation_narrow}
\end{figure}

We conduct an ablation study to isolate how data scale and model size affect performance. We vary each factor independently while keeping the model architecture and training protocol fixed.

\subsubsection{Effect of Training Data Size}

\noindent\textbf{Setup.}
To evaluate how the data scale affects our method, we fine-tuned judge-specific models on subsets of their available training data (25\%, 50\%, 75\%, and 100\%), while fixing the LoRA rank at 8. The 0\% point represents a non-personalized base model. 
The full dataset ranges from 2,500 to 3,100 examples per judge, which corresponds to $\sim$ 310k--375k tokens. 

\noindent\textbf{Findings.}
We observe a two-regime learning curve as the amount of training data increases.
With $\leq$25\% of the data ($\sim$650 examples), the personalized model remains close to the base model, showing limited improvements over the baseline.
Between 25\% and 50\% ($\sim$650$\to$1{,}500 examples), both semantic fidelity (BERTScore) and stylistic alignment (JSD) improve sharply, after which gains largely plateau.

\subsubsection{Effect of LoRA Rank}

\textbf{Setup.}
To determine the minimal parameter tradeoffs for effective adaptation, we varied the LoRA rank $r \in \{2, 4, 8, 16, 32\}$ on the full dataset, which is proportional to model size. 

\noindent\textbf{Findings.}
LoRA rank exhibits diminishing returns. The largest improvement occurs at $r=2$: for Judge A, BLEU increases from 0.9 (Vanilla) to 14.9, reaching $\sim$75\% of the best observed performance.
Across judges, semantic fidelity (BERTScore) and stylistic alignment (JSD) improve mainly at low ranks and largely plateau by $r=4$.
BLEU continues to vary more strongly across judges and tends to benefit from larger $r$, but the average gains beyond $r=8$ are small.

\subsubsection{Data Size vs. Model Size}
To disentangle the effect of training volume from model size, we compare (i) doubling the amount of judge-specific training data (e.g., $50\% \to 100\%$) to (ii) doubling the LoRA rank ($r \to 2r$), holding the remaining setup fixed.
Doubling the data produces substantially larger gains: on average, \textbf{+2.68 BLEU}, along with consistent improvements in semantic fidelity (BERTScore increases of $\sim$0.02) and in our style metrics.
In contrast, doubling the rank yields only a modest \textbf{+0.77 BLEU} increase and negligible changes in semantic and style scores.

These results indicate diminishing returns from increasing adapter rank, while additional training examples continue to improve both content preservation and stylistic alignment.
In particular, our style divergence measure (JSD) plateaus early in the rank sweep (around $r{=}4$), but continues to improve as more judge text is added, suggesting that personalization quality is more constrained by data coverage than by adapter capacity.


\subsection{Qualitative Evaluation}

To complement the automatic metrics, we conducted a blind human evaluation of 100 question-answer instances. A human annotator with basic legal training, blinded to answer origin, evaluated each answer for responsiveness, logical coherence and legal plausibility, and resemblance to authentic judicial writing. All ground-truth answers (40/40) were labelled as addressing the question, logically coherent, and resembling judicial writings. Model-generated answers were rated positively in 71.2\% of cases for answering the question, 82.7\% for coherence and legal plausibility, and 80.8\% for judicial authenticity. These results suggest that, despite some failure cases, most generated answers are sound and stylistically judge-like.

In addition, we qualitatively examined the answers of the base model, CLM, and CoLA on a random sample of identical questions. We observed a consistent pattern: the base model often produced long, generic responses resembling encyclopedic explanations. CLM reduced this behavior, but remained verbose and occasionally unfocused. In contrast, CoLA generated more concise outputs with a more authentic judicial tone. 

However, CoLA also exhibits several failure modes, including overly short and uninformative answers, occasional drift into irrelevant legal text, and, in some cases, repetition. We also observe instances where CoLA maintains stylistic fidelity but misses key legal reasoning present in the ground truth. Representative examples are provided in Appendix~\ref{sec:error_analysis}.

\subsection{Robustness to Catastrophic Forgetting}

To assess potential catastrophic forgetting, we evaluated the personalized models on three out-of-domain settings: (i) general Hebrew text from Wikipedia, (ii) Hebrew legal text from the Israeli Supreme Court written by judges excluded from training, using the Israeli Supreme Court Dataset~\cite{muchnik2023israeli_supreme_court_dataset}, and (iii) English commonsense reasoning using HellaSwag~\cite{zellers2019hellaswag}. We report perplexity for the Hebrew evaluations and multiple-choice accuracy for HellaSwag. 

Relative to the base model, CLM significantly improves perplexity in both Hebrew settings across all 29 judges. CoLA shows no significant difference on Hebrew Wikipedia, but significantly improves perplexity on the out-of-domain Supreme Court text. On HellaSwag, CLM slightly improves over the base model, while CoLA shows a small decrease (66.4\% for the base model, 67.3\% for CLM, and 64.9\% for CoLA; all $p<0.05$). Overall, these findings suggest no catastrophic forgetting, but at most a modest tradeoff in English general knowledge.

\subsection{Next-Step Reasoning Prediction}

As an additional probe, we evaluated the models on a multiple-choice next-step reasoning task. Given one to three reasoning sentences from a judicial decision, each model was asked to identify the correct subsequent reasoning sentence from a set of candidate continuations drawn from the same case. Candidate answers were scored by their conditional likelihood under the model given the preceding context, and the highest-scoring continuation was selected. We found that none of the evaluated models, including the base model and CoLA, achieved strong performance on this task. This suggests that predicting the next reasoning step remains a difficult task, even for personalized models, that requires further supervision.

\section{Discussion}
Our findings show that reasoning signatures can be effectively captured even in low-resource, morphologically rich settings. 
This result challenges the prevailing perception that training reasoning capabilities requires massive computing resources and web-scale datasets.
By successfully modeling distinct judicial decisions with limited compute (a 4B parameter model with LoRa) and a limited amount of data, we show that the ``cognitive fingerprint'' of a decision-maker is compressible.
Critically, this efficiency is achieved by our synthetic-organic alignment pipeline.
Exposure to the raw corpora (via CLM) was found to be less efficient, most likely due to the dilution of reasoning signals among other types of information.
However, when the same data was structured into synthetic instruction pairs, the model was able to learn judges' reasoning patterns with far fewer tokens.
This suggests that the primary bottleneck in personalization is not the quantity of user data, but rather the structure of supervision signal, making personalization potentially viable in other domains where there are similar amounts of reasoning traces exist. 

Beyond these efficiency gains, the study also sheds light on the theoretical dichotomy between surface-level style and deep reasoning, a critical tension in current alignment research~\cite{krishna-etal-2020-reformulating,neelakanteswara-etal-2024-rags,xu-etal-2025-personalized}.
The contrast between our RAG and parameter-adapted (CoLA) models is illustrative: While RAG effectively mimicked the syntactic style of the domain (improving JSD-POS scores), it failed to capture the semantic reasoning needed to address substantive questions. 
Conversely, parameter-tuning excelled at emulating judges' reasoning but showed slightly less adherence to surface-level part-of-speech distributions.
This suggests that a ``persona'' may be decomposed more effectively into two layers: surface-level stylistic patterns, which retrieval can capture, and deeper reasoning priors, which require parametric adaptation. 
Future research may explore techniques for optimizing both layers simultaneously. 

\section{Conclusion}
In this work, we presented a framework for personalizing LLMs to the reasoning styles of individual judges.
By converting unstructured judicial reasoning into synthetic-organic instruction-tuning data, we introduced supervision signals that enabled efficient training. 
Our results show that CLM followed by instruction-tuning significantly outperforms all other baselines, including a state-of-the-art reasoning model with RAG and models trained on considerably more data, all while remaining computationally feasible and effective in a low-resource, morphologically rich language like Hebrew. 
We also demonstrated consistent improvements across lexical, stylistic, and semantic dimensions, ultimately making LLM's simulated output indistinguishable from the reasoning of the real judge. 
These results highlight the potential to efficiently train personalized reasoning models in low-resource settings, which future work could explore in other expert domains (e.g., medicine) and more complex reasoning tasks (e.g., tasks that require planning). 


\section{Limitations}
The current study has several important limitations. 
First, our framework focuses on the reasoning and style of granular decisions dispersed throughout a verdict. 
While these decisions surely affect the final ruling, our approach does not model case-level reasoning, which is inherently sparse, or account for shifts in reasoning over time. 
Moreover, future work could also examine ways to integrate existing knowledge (e.g., the fact of a case) in reasoning supervision signals. 

Second, our personalization objective prioritizes emulating the voice and reasoning signature of a judge rather than ensuring strict legal correctness, which we considered outside the scope of the current work. 
However, factuality of personalization models, especially in the legal domain, is a worthy avenue of future research, which should probably focus on high-resource languages with a robust knowledge base. 
Finally, the current study focused on a single low-resource language and just one legal system. It is important to examine how these findings generalize to other settings. 

\section{Ethical considerations}
As this research relies exclusively on the analysis of existing, publicly available court records generated by public officials in their official capacity and intended for public scrutiny, involving no interaction or intervention, it does not constitute human subjects research requiring institutional review.
Nevertheless, we recognize that the sensitivity of the domain and the focus on personalization require further considerations beyond standard regulatory compliance.
Here, we discuss these additional considerations, namely the potential for misuse, risks of bias and amplification, privacy, and our mitigation strategies.

\textbf{Potential for misuse:} The development of models capable of emulating specific judges raises potential concerns regarding the integrity of the judicial process. 
Chiefly, there is a risk that such tools could be used by litigants to ``game the system'', tailoring arguments to exploit a judge's predicted linguistic or philosophical preferences. 
We note, however, that adapting arguments to the perceived judicial philosophy of a judge is already a common, albeit manual, practice in litigation strategy~\cite{posner2008judges}. 
Critically, one must distinguish between the \textit{emulation of reasoning patterns} and the exercise of \textit{judicial discretion}. 
While our models can capture some persistent patterns of reasoning of specific judges, they inherently lack the moral agency and accountability that characterize human decision. 
Using these tools to predict future outcomes is fundamentally flawed: they may replicate previous pattern, but are far from able to weigh the merits of future cases.

\textbf{Bias amplification and fairness:} Judicial decisions inherently reflect historical, systemic, and individual biases present in the legal system. 
By optimizing for high-fidelity emulation, personalized models risk not only reproducing but also perpetuating and potentially amplifying biases, under the pretense of mathematical neutrality. 
Moreover, similar to general-purpose LLMs, users may conflate linguistic fluency with correctness or objectivity. 
We therefore explicitly state that these models are designed to capture the idiosyncrasies of specific reasoning signatures, including their flaws and biases, and should be thoroughly audited for fairness before any potential deployment in decision-support contexts.

\textbf{Data privacy and mitigation:} To address these risks and minimize potential harm, we have implemented strict protocols regarding data handling and model distribution. 
Although our training data consists entirely of public domain verdicts, we recognize that generative models can hallucinate plausible but false information, potentially placing individuals in unanticipated circumstances. 
This applies both to the individuals mentioned in the corpus and to the judges themselves. 
Therefore, in compliance with principles of responsible AI data minimization, we remove identifying information about judges and avoid analyzing or reporting findings associated with specific identities of the judges analyzed. 
Furthermore, we refrain from releasing the trained judge-specific model checkpoints, which serve as a safeguard against potential misuse, production of hallucinated content, and the malicious deployment of these personas in real-world settings.

\section{Acknowledgments}
This research was supported by the Israeli Innovation Authority, grant number 81686. We thank members of our research team for helpful comments and discussions.


\appendix

\section{Appendix}
\label{sec:appendix}

\subsection{Next Token Prediction - CLM Baseline Comparison Table}

\begin{table*}
\centering
\small
\setlength{\tabcolsep}{3pt}
\begin{tabular}{lcccccc}
\toprule
Model & BLEU $\uparrow$ & BERTScore-F1 $\uparrow$ & ROUGE-1 $\uparrow$ & ROUGE-2 $\uparrow$ & ROUGE-L $\uparrow$ & POS-JSD $\downarrow$ \\
\midrule
\multicolumn{7}{l}{\textit{Non-personalized baselines}} \\
Vanilla-DictaLM3        & 5.533 & 0.076 & 0.073 & 0.021 & 0.063 & -0.042 \\
Vanilla-Gemma (1B)      & 6.010 & 0.124 & 0.101 & 0.025 & 0.087 & -0.040 \\
Vanilla-Gemma (4B)      & 5.597 & 0.079 & 0.081 & 0.023 & 0.071 & -0.013 \\
Mixed-Gemma (1B)        & 4.662 & 0.065 & 0.067 & 0.021 & 0.057 & -0.042 \\
Mixed-Gemma (4B)        & 4.383 & 0.042 & 0.050 & 0.019 & 0.042 & -0.019 \\
Joint-Gemma (1B)        & 2.127 & 0.016 & 0.013 & 0.006$^{*}$ & 0.013 & -0.005$^{*}$ \\
Joint-Gemma (4B)        & 2.349 & 0.016 & 0.017 & 0.008 & 0.015 & -0.010 \\
\addlinespace
\multicolumn{7}{l}{\textit{Alternative personalization paradigms}} \\
Pers-CLM (1B)           & 0.643 & -0.062 & 0.006$^{*}$ & 0.002$^{*}$ & 0.006$^{*}$ & -0.002$^{*}$ \\
Pers-IT (1B)            & 6.529 & 0.142 & 0.106 & 0.026 & 0.092 & -0.100 \\
Pers-IT (4B)            & 5.542 & 0.066 & 0.084 & 0.022 & 0.072 & -0.033 \\
Chain-of-LoRA (4B)      & 5.101 & 0.062 & 0.074 & 0.018 & 0.063 & -0.038 \\
\bottomrule
\end{tabular}%
\caption{Comparison to Pers-CLM (4B) using per-judge paired deltas (method $-$ Pers-CLM 4B), averaged over judges. We report two-sided Wilcoxon signed-rank tests over judges for each metric; all entries are significant at $p<0.05$ unless marked (*).}
\label{tab:clm_vs_all_models_appendix}
\end{table*}

Table~\ref{tab:clm_vs_all_models_appendix} reports the full CLM improvement (absolute difference) of the best-performing model relative to the other methods.

\subsection{Impact of Model Size}
Table~\ref{tab:4b_vs_1b_wilcoxon} compares the performance of 1B and 4B model variants. We find that increasing model size consistently improves performance. The 4B models achieve higher mean scores across nearly all metrics in both instruction and raw-text settings.

\begin{table*}
\centering
\small
\setlength{\tabcolsep}{3pt}
\begin{tabular}{lcccccc}
\toprule
Method & BLEU $\uparrow$ & BS-F $\uparrow$ & POS $\downarrow$ & R-1 $\uparrow$ & R-2 $\uparrow$ & R-L $\uparrow$ \\
\midrule

\multicolumn{7}{l}{\textit{Next Token Prediction (CLM) setting}} \\
CLM     & \textbf{6.690} / 6.047 & 0.606 / \textbf{0.667} & \textbf{0.059}$^{*}$ / 0.061$^{*}$ & \textbf{0.123}$^{*}$ / 0.117$^{*}$ & \textbf{0.027}$^{*}$ / 0.025$^{*}$ & \textbf{0.108}$^{*}$ / 0.102$^{*}$ \\
IT      & \textbf{1.148} / 0.161 & \textbf{0.540} / 0.463 & \textbf{0.092} / 0.159 & \textbf{0.040} / 0.017 & \textbf{0.005} / 0.001 & \textbf{0.036} / 0.015 \\
Joint   & 4.341$^{*}$ / \textbf{4.563}$^{*}$ & \textbf{0.589}$^{*}$ / \textbf{0.589}$^{*}$ & 0.069 / \textbf{0.064} & 0.107$^{*}$ / \textbf{0.110}$^{*}$ & 0.019$^{*}$ / \textbf{0.022}$^{*}$ & 0.092$^{*}$ / \textbf{0.094}$^{*}$ \\
Mixed   & \textbf{2.308} / 2.028 & \textbf{0.563} / 0.541 & \textbf{0.078} / 0.101 & \textbf{0.073} / 0.056 & \textbf{0.008} / 0.006 & \textbf{0.066} / 0.051 \\
Vanilla & \textbf{1.093} / 0.680 & \textbf{0.527} / 0.482 & \textbf{0.072} / 0.100 & \textbf{0.042} / 0.023 & \textbf{0.004} / 0.002 & \textbf{0.037} / 0.020 \\
\addlinespace
\multicolumn{7}{l}{\textit{Instruction setting}} \\
CLM     & \textbf{1.565} / 1.040 & \textbf{0.561} / 0.534 & \textbf{0.073} / 0.077 & \textbf{0.201} / 0.183 & \textbf{0.060} / 0.044 & \textbf{0.136} / 0.122 \\
CoLA    & \textbf{13.313} / 9.948 & \textbf{0.639} / 0.597 & \textbf{0.073} / 0.079 & \textbf{0.385} / 0.332 & \textbf{0.200} / 0.147 & \textbf{0.317} / 0.268 \\
IT      & 8.339$^{*}$ / \textbf{10.032}$^{*}$ & \textbf{0.619} / 0.604 & \textbf{0.074} / 0.077 & \textbf{0.349}$^{*}$ / 0.345$^{*}$ & 0.151$^{*}$ / \textbf{0.155}$^{*}$ & 0.274$^{*}$ / \textbf{0.276}$^{*}$ \\
Joint   & \textbf{4.401} / 3.278 & \textbf{0.602} / 0.570 & \textbf{0.082} / 0.086 & \textbf{0.318} / 0.286 & 0.111 / 0.083 & 0.239 / 0.211 \\
Mixed   & \textbf{4.108} / 3.302 &\textbf{0.596} / 0.568 & \textbf{0.085} / 0.088 & \textbf{0.316} / 0.289 & \textbf{0.107} / 0.084 & \textbf{0.234} / 0.214 \\
Vanilla & \textbf{1.316}$^{*}$ / 1.309$^{*}$ & \textbf{0.526} / 0.505 & \textbf{0.090} / 0.099 & \textbf{0.181} / 0.178 & \textbf{0.050} / 0.049 & \textbf{0.124} / \textbf{0.124} \\

\bottomrule
\end{tabular}%
\caption{Model size comparison (mean across judges). Each cell reports \textit{4B / 1B}. Asterisks (*) mark entries where the 4B and 1B means are not significantly different under a paired Wilcoxon signed-rank test ($p \ge 0.05$). Results are shown separately for the instruction setting and the Next Token Prediction (CLM)  setting.}
\label{tab:4b_vs_1b_wilcoxon}
\end{table*}

\subsection{Cross-Judge Specificity: Full Results}
Table~\ref{tab:cross_judge_all_methods} reports the cross-judge specificity summary for all methods.
Each entry shows the mean test-set centered gap across judges, followed by the percentage of judges for which the paired gap is statistically significant.
Consistent with Figure~\ref{fig:forest-onecol}, gaps are predominantly positive for lexical and semantic metrics (BLEU, ROUGE, and BERTScore), indicating that judge-matched models tend to score higher than models trained on other judges.
By contrast, the structural JSD-POS gaps are smaller and less consistently significant across judges, suggesting weaker judge-specific separation under POS-distribution similarity.

\begin{table*}
\centering
\small
\setlength{\tabcolsep}{3pt}
\begin{tabular}{lcccccc}
\toprule
Method & BLEU $\uparrow$ & BS-F $\uparrow$ & POS $\downarrow$ & R-1 $\uparrow$ & R-2 $\uparrow$ & R-L $\uparrow$ \\
\midrule
\multicolumn{7}{l}{\textit{Next Token Prediction (CLM) Setting}} \\
CLM-1B & 4.523 (0.93) & 0.031 (0.76) & 0.012 (0.59) & \textbf{0.039 (0.66)} & 0.018 (0.59) & \textbf{0.036 (0.69)} \\
CLM-4B & \textbf{4.925 (0.86)} & \textbf{0.046 (0.83)} & 0.011 (0.72) & 0.037 (0.59) & \textbf{0.020 (0.52)} & 0.035 (0.62) \\
CoLA-4B & 0.810 (0.55) & 0.025 (0.52) & 0.007 (0.28) & 0.014 (0.34) & 0.006 (0.21) & 0.013 (0.34) \\
IT-1B & 0.026 (0.28) & 0.008 (0.28) & 0.003 (0.28) & 0.002 (0.14) & 0.000 (0.41) & 0.002 (0.10) \\
IT-4B & 0.364 (0.66) & 0.012 (0.52) & \textbf{0.002 (0.45)} & 0.004 (0.03) & 0.002 (0.14) & 0.004 (0.03) \\
\addlinespace
\multicolumn{7}{l}{\textit{Instruction Setting}} \\
CLM-1B & 0.226 (0.79) & 0.008 (0.76) & \textbf{0.001 (0.41)} & 0.007 (0.34) & 0.008 (0.79) & 0.004 (0.48) \\
CLM-4B & 0.424 (0.79) & 0.012 (0.69) & 0.003 (0.59) & 0.010 (0.52) & 0.011 (0.86) & 0.008 (0.59) \\
CoLA-1B & 7.600 (0.83) & 0.051 (0.83) & 0.012 (0.69) & 0.079 (0.83) & 0.090 (0.83) & 0.081 (0.86) \\
CoLA-4B & \textbf{10.062 (0.93)} & \textbf{0.061 (0.93)} & 0.013 (0.66) & \textbf{0.099 (0.97)} & \textbf{0.117 (0.97)} & \textbf{0.105 (0.93)} \\
IT-1B & 7.285 (0.83) & 0.050 (0.83) & 0.010 (0.59) & 0.075 (0.86) & 0.086 (0.86) & 0.078 (0.86) \\
IT-4B & 5.098 (0.79) & 0.040 (0.86) & 0.009 (0.62) & 0.060 (0.83) & 0.067 (0.83) & 0.062 (0.86) \\
RAG-3k & 1.125 (0.90) & 0.023 (0.90) & 0.003 (0.21) & 0.024 (0.90) & 0.025 (0.90) & 0.022 (0.90) \\
RAG-5k & 1.286 (0.93) & 0.025 (0.93) & 0.003 (0.34) & 0.026 (0.90) & 0.028 (0.93) & 0.023 (0.90) \\
\bottomrule
\end{tabular}%
\caption{Cross-judge specificity across methods. For each judge $j$, we evaluate the matched model $M_j$ on $T_j$ and compare it to models trained on other judges, evaluated on the same $T_j$. Each cell reports the mean centered gap $\Delta_j$ (averaged over judges); the value in parentheses is the fraction of judges where the gap is significant under bootstrap. Positive gaps are better.}
\label{tab:cross_judge_all_methods}
\end{table*}

\subsection{Training Setup}
\label{sec:appendix_training}

All experiments were run on a single NVIDIA RTX~4090 GPU. To handle memory constraints, we used a dynamic gradient accumulation strategy that adjusted the number of accumulation steps as needed to maintain training stability. All other hyperparameters are listed in \autoref{tab:hyperparameters}.

\onecolumn
\begin{table*}[t] 
    \centering
    \small

    \begin{minipage}[t]{0.48\textwidth}
        \centering
        \begin{tabular}{l r}
            \toprule
            \textbf{Hyperparameter} & \textbf{Value} \\
            \midrule
            \multicolumn{2}{l}{\textit{LoRA Configuration}} \\
            Rank ($r$) & 16 \\
            Alpha ($\alpha$) & 32 \\
            Dropout & 0.05 \\
            Target Modules & All linear layers* \\
            \midrule
            \multicolumn{2}{l}{\textit{Training Dynamics}} \\
            Learning Rate & $2.0 \times 10^{-4}$ \\
            Optimizer & Paged AdamW 8-bit \\
            Scheduler & Linear \\
            Warmup Steps & 5 \\
            Max Steps & 2,500 \\
            Max Sequence Length & 512 \\
            Weight Decay & 0.01 \\
            Seed & 3407 \\
            \bottomrule
        \end{tabular}
        \caption{Hyperparameters used for fine-tuning. * Target modules include $q, k, v, o$ projections, gates, and the LM head.}
        \label{tab:hyperparameters}
    \end{minipage}
    \hfill 
    \begin{minipage}[t]{0.48\textwidth}
        \centering
        \begin{tabular}{lc}
            \toprule
            \textbf{Method} & \textbf{Acc. (\%)} \\
            \midrule
            Random Baseline & 50.0 \\
            \addlinespace
            \multicolumn{2}{l}{\textbf{Non-personalized}} \\
            \hspace{3mm}Vanilla-1B & 80.6$^{*}$ \\
            \hspace{3mm}Vanilla-4B & 80.6$^{*}$ \\
            \hspace{3mm}Dicta-3 & 81.7$^{*}$ \\
            \hspace{3mm}Mixed-1B & 84.0$^{*}$ \\
            \hspace{3mm}Mixed-4B & 84.0$^{*}$ \\
            \hspace{3mm}Joint-1B & 84.1$^{*}$ \\
            \hspace{3mm}Joint-4B & 82.5$^{*}$ \\
            \addlinespace
            \multicolumn{2}{l}{\textbf{Alternative methods}} \\
            \hspace{3mm}IT-4B & 63.4$^{*}$ \\
            \hspace{3mm}IT-1B & 68.1$^{*}$ \\
            \hspace{3mm}CoLA-4B & 61.8$^{*}$ \\
            \hspace{3mm}CLM-4B & 56.0$^{*}$ \\
            \hspace{3mm}CLM-1B & 56.5$^{*}$ \\
            \addlinespace
            \multicolumn{2}{l}{\textbf{Reference}} \\
            \hspace{3mm}Ground Truth & 94.9$^{*}$ \\
            \bottomrule
        \end{tabular}
        \caption{Authorship accuracy relative to ground truth. $*$ denotes $p<0.05$.}
        \label{tab:acc_vs_random_appendix}
    \end{minipage}
\end{table*}

\subsection{Question-Answer Generation}
\label{sec:qa-gen}
For that task, we used an agentic workflow implemented entirely with OpenAI's GPT~4.1-mini and GPT~4o-mini, applying the steps:

\begin{enumerate}[nosep, leftmargin=*]
    \item \textbf{Reasoning Sentence Extraction} We process each verdict with a chain-of-thought prompt to identify sentences that express the judge's legal reasoning.
    \item \textbf{Judicial Reasoning Validation} We pass the extracted flagged sentences to a separate validator prompt that confirms they indeed contain judicial reasoning; sentences that fail validation are removed.
    \item \textbf{Question Generation} Based on the extracted reasoning sentences, we generate questions whose answers refer to the reasoning sentences.
    \item \textbf{Validation} We then verify that the questions match the answers; if validation fails twice, the pair is discarded.
\end{enumerate}

The exact prompts used in each stage of the workflow are provided in Section~\ref{sec:qa-prompts}.


\subsection{Qualitative Error Analysis}
\label{sec:error_analysis}

Below we present representative question--answer pairs comparing the outputs of the Base model, CLM, and CoLA on identical inputs.

\begin{longtable}{@{} p{0.10\textwidth} p{0.86\textwidth} @{}}
\toprule
\textbf{Model} & \textbf{Output} \\
\midrule
\endfirsthead
\toprule
\textbf{Model} & \textbf{Output} \\
\midrule
\endhead
\midrule
\multicolumn{2}{r}{\footnotesize\itshape Continued on next page} \\
\bottomrule
\endfoot
\bottomrule
\endlastfoot

\textbf{Question} &
\foreignlanguage{hebrew}{מהו מתחם העונש ההולם שנקבע בנסיבות ביצוע העבירה?} \\
\midrule
Ground Truth &
\foreignlanguage{hebrew}{נוכח המפורט לעיל, איני סבורה כי מתחם העונש שולל קצה נמוך של מאסר על תנאי בלבד, כפי שבקשה המאשימה לקבוע, וסבורני כי מתחם העונש ההולם בנסיבות ביצוע העבירה נע בין מאסר על-תנאי בצירוף ענישה כספית, ובין שבעה חודשי מאסר בפועל, שיכול וירוצו בעבודות שירות.} \\
Base &
\foreignlanguage{hebrew}{מתחם העונש ההולם (Range of Sentencing Options) שנקבע בנסיבות ביצוע העבירה הוא סט המינימום והמקסימום של עונשים שהמדינה יכולה לגזור על נאשם שהורשע בעבירה. הוא לא מחייב את בית המשפט לגזור את העונש הנמוך ביותר או הגבוה ביותר, אלא מספק לו טווח אפשרויות. הגדרת מתחם העונש ההולם מורכבת ומושפעת ממספר גורמים, ביניהם: 1. חוק העבירה: החוק עצמו מגדיר את סוגי העונשים האפשריים [...] 2. סוג העבירה: חומרה של העבירה היא גורם מרכזי. עבירות חמורות יותר (כגון [...]} \\
CLM &
\foreignlanguage{hebrew}{אכן, שיתוף הפסילה המקילה והארוכה הנה התייחסות לחומרת העבירה שביצע הנאשם (פסיקה שניתנה בסגור או בפני שופטת). יחד עם זאת, בשני מקרים שהוצעו, הנאשם לא ביצע את העבירות המיוחסות לו, והם אינם מלמדים על מתחם ענישה הולם או מתאים. לכל מקרה ייתכן ויש ליתן ביטוי שונה בחומרת המתחם [...]} \\
CoLA &
\foreignlanguage{hebrew}{בנסיבות ביצוע העבירה, בשקלול הנסיבות שפורטו לעיל, אני קובעת כי מתחם העונש ההולם בנסיבות ביצוע העבירה נע בין 14 ל-30 חודשי מאסר.} \\
\midrule
 &
The Base model provides a generic, encyclopedic definition of sentencing ranges with numbered bullet points. CLM produces legal-sounding but unfocused text. CoLA generates a concise, first-person judicial ruling that mirrors the ground truth's structure. \\

\midrule\addlinespace[0.5em]

\textbf{Question} &
\foreignlanguage{hebrew}{מהו האיזון בין השיקולים שהנחה את בית המשפט בגזירת העונש על הנאשם?} \\
\midrule
Ground Truth &
\foreignlanguage{hebrew}{באיזון בין השיקולים השונים, סבורני כי יש לגזור על הנאשם עונש בחלקו המרכזי המתחם.} \\
Base &
\foreignlanguage{hebrew}{בגזירת עונש על נאשם, בית המשפט צריך לאזן בין מגוון רחב של שיקולים, שמטרתם היא להעניש את הנאשם בצורה הולמת, למזער את הנזקים ולתרום להסדרה ולשיקום. הנה פירוט של השיקולים העיקריים שהנחו אותו: 1. ענישה: חומרת העבירה: זהו השיקול הבסיסי ביותר [...] נסיבות העבירה [...] עקרונות ענישה [...]} \\
CLM &
\foreignlanguage{hebrew}{בעניין זה יש לבחון את הערכים החברתיים אשר נפגעו עקב ביצוע העבירה, ואת מידת הפגיעה בערך המוגן [...] בגזירת העונש המתאים לנאשם, בגדרי מתחם העונש ההולם, יש להתחשב בנסיבות שאינן קשורות בביצוע העבירה. במסגרת זו מן הראוי לתת את הדעת לכך שהנאשם צעיר [...]} \\
CoLA &
\foreignlanguage{hebrew}{באיזון בין השיקולים השונים, יש לכבד את המלצת שירות המבחן בעניינו של הנאשם אשר סבורים כי העונש מנוטר והגורם לעונש של עבודות שירות, במקרים שבהם קיימת תמטיקה של חלוף דיינו וכי עונש מאסר בכליאה יפגע בו.} \\
\midrule
 &
The Base model produces a textbook-style explanation. CLM gives a verbose legal discussion. CoLA opens with the same phrasing as the ground truth and produces a case-specific judicial statement, though some phrasing is imprecise. \\

\midrule\addlinespace[0.5em]

\textbf{Question} &
\foreignlanguage{hebrew}{כיצד הודאת הנאשם בעובדות כתב האישום משפיעה על ההליך השיפוטי?} \\
\midrule
Ground Truth &
\foreignlanguage{hebrew}{לכל האמור לעיל יש להוסיף את הודאתו של הנאשם בעובדות כתב האישום, המבטאת נטילת אחריות מצדו על מעשיו ואשר הובילה לחיסכון בזמן שיפוטי יקר.} \\
Base &
\foreignlanguage{hebrew}{התנגדות לנאשם בעובדות כתב האישום משפיעה באופן משמעותי על ההליך השיפוטי, ויש לה השלכות משפטיות מורכבות. הנה פירוט של האופן שבו זה מתבטא: 1. תחילת ההליך: הגשת התנגדות: הנאשם יכול להגיש התנגדות [...] 2. חקירה וניהול הדיון: חקירת התנגדות [...]} \\
CLM &
\foreignlanguage{hebrew}{הודאת הנאשם בעובדות כתב האישום, מקלה על תפקידו של בית המשפט במניעת הצורך לחקור את הנסיבות שבעבירות נשוא כתב האישום, ובכך חוסכת את זמן שיקול הדעת והזמן שירוצה על ידו [...]} \\
CoLA &
\foreignlanguage{hebrew}{הודאה זו החזירה את הביצועים מהרבה יותר ממה שהיה צפוי.} \\
\midrule
&
CoLA produces an extremely short, uninformative sentence that is meaningless in this legal context. Both the Base model and CLM, despite their verbosity, at least convey relevant legal content. This illustrates the failure mode of overly terse output. \\

\midrule\addlinespace[0.5em]

\textbf{Question} &
\foreignlanguage{hebrew}{מהי המדיניות הכללית בענישה בעבירות הצתה של בתי מגורים?} \\
\midrule
Ground Truth &
\foreignlanguage{hebrew}{מדיניות הענישה הנהוגה בעבירות הצתה של בתי מגורים, הינה של חומרה.} \\
Base &
\foreignlanguage{hebrew}{העונש בעבירות הצתה של בתי מגורים הוא עונש חמור מאוד, והמדיניות הכללית בענישה מתייחסת למגוון גורמים והסדרה של רמות ענישה, המותאמות למאפיינים של המקרה הספציפי. הנה סקירה כללית: 1. סיווג העבירה [...]} \\
CLM &
\foreignlanguage{hebrew}{המשפט העליון ציין שטיב העונש הראוי בעבירות הצתה צריך לתוא את חומרת מעשה ההצתה – ככל שהוא גורם לנזק גדול יותר, גם הענישה צריכה להיות חמורה יותר. שכן הנזק שנגרם כתוצאה מהצתת הדירה הוא חמור וממשי [...]} \\
CoLA &
\foreignlanguage{hebrew}{כל בר כשר המשנה את פני השמש ביום האש, יפנה מלכתחילה לעבר התחרויות המדגישות את נצחיות הזיכרון ואת מעלתה המיוחדת של הדרישת הדין להרתעה.} \\
\midrule
&
CoLA drifts into poetic, irrelevant language that bears no resemblance to legal reasoning. While the Base and CLM answers are verbose, they at least address sentencing policy for arson. This exemplifies the failure mode of drifting into incoherent text. \\

\midrule\addlinespace[0.5em]

\textbf{Question} &
\foreignlanguage{hebrew}{מהו הכיעור המיוחד הקשור בעבירת הסחיטה באיומים?} \\
\midrule
Ground Truth &
\foreignlanguage{hebrew}{העבירה של סחיטה באיומים טבועה בכיעור מיוחד, מעבר לכיעור הרגיל הכרוך במעשים פליליים.} \\
Base &
\foreignlanguage{hebrew}{הכיעור המיוחד הקשור בעבירת הסחיטה באיומים הוא האיום. האיום, בניגוד לאיום רגיל (שאינו גורם עונתי), הוא תכונה ייחודית של עבירת הסחיטה, והוא מהווה את הבסיס לעונש החמור יותר [...] סחיטה: עבירת הסחיטה מתרחשת כאשר אדם מפעיל כוח או איומים [...]} \\
CLM &
\foreignlanguage{hebrew}{הכיעור המיוחד הקשור בעבירת הסחיטה באיומים, מתבטא בהשפעה הפסיכולוגית הנובעת מהסכנה שהיא טומנת בחובה, שהיא משמעותית יותר מזו הקשורה לעבירות רכוש בנסיבות כלליות [...]} \\
CoLA &
\foreignlanguage{hebrew}{העבירה של סחיטה באיומים טומנת בחובה כיעור מיוחד, הנוגע לאינטרסים חברתיים בעלי חשיבות קלאסית.} \\
\midrule
 &
CoLA closely echoes the ground truth's phrasing and judicial tone. \\

\midrule\addlinespace[0.5em]

\textbf{Question} &
\foreignlanguage{hebrew}{מדוע יש לראות בעבירות מס כעבירות גזל כספי הציבור ולא רק כעבירות רגולטוריות?} \\
\midrule
Ground Truth &
\foreignlanguage{hebrew}{הפסיקה חוזרת ומדגישה כי אין לראות בעבירות מס משום עבירות רגולטוריות גרידא, שכן משמעותן המעשית היא גזל כספי הציבור.} \\
Base &
\foreignlanguage{hebrew}{השאלה מדוע ראוי לראות בעבירות מס כעבירות גזל כספי הציבור, ולא רק כעבירות רגולטוריות, היא שאלה מורכבת עם היבטים משפטיים, אתיים וכלכליים. הנה מספר טיעונים התומכים בגישה זו: 1. פגיעה חמורה בנכסי המדינה: המס הוא אמצעי מימון [...] הונאה על חשבון המטרה [...]} \\
CLM &
\foreignlanguage{hebrew}{שאלה מצוינת. כפי שציינת, בתי המשפט שונים – ובוודאי שגם בפסיקה העולה מהרשות העוסקת בכך (ר' למשל רע"פ 7009/20 יהודה ואח' נ' מדינת ישראל [פורסם בנבו] (2021) – בעניין עבירות המס נדרשת גישה מהותית, השוות לזו של גזל [...]} \\
CoLA &
\foreignlanguage{hebrew}{כמעט למותר לציין כי בעבירות מס, כבעבירות של הגבלת יסודות חברה ועבירות גניבה, אי אפשר לראותן משום עבירות רגולטוריות גרידא.} \\
\midrule
 &
 The Base model provides a lengthy essay, while CLM fabricates a case citation. CoLA is closest in style but misses the substantive argument. \\

\end{longtable}

\subsection{Question--Answer Generation Prompts}
\label{sec:qa-prompts}

Below we provide the exact prompts used in each step of the question--answer generation pipeline.

\subsubsection*{Step 1: Reasoning Sentence Extraction}
\textbf{Model:} GPT-4.1-mini

\begin{quote}
\small
\foreignlanguage{hebrew}{%
\textbf{\# תפקיד המערכת}

אתה אנליסט משפטי. שלוף מתוך פסק הדין אך ורק משפטים הסברתיים — כלומר, משפטים שבהם השופט מסביר את שיקוליו, נימוקיו או עקרונות שבהם הסתמך.
החזר את כל המשפטים המלאים, ללא שינוי ניסוח או קיצור.
החזר את הפלט במבנה JSON של רשימת מחרוזות תחת השם `משפטים'.
אל תסווג או תדרג את המשפטים, רק תאסוף אותם.
אם אין משפטים הסברתיים — החזר רשימה ריקה.

\textbf{\# הנחיות כלליות}
\begin{enumerate}
\item החזר את המשפטים בדיוק כפי שהם כתובים בפסק הדין — ללא קיצור, סיכום, שינוי ניסוח או שלוש נקודות (...).
\item אל תכלול משפטים טענתיים או כלליים בלבד. החזר רק משפטים שמכילים נימוק, סיבה או עקרון ברור.
\item אל תכלול משפטים שנאמרו ע"י ב"כ המאשימה או ב"כ הנאשם, רק משפטים הסברתיים שהשופט הסביר את שיקוליו.
\item אל תכלול משפטים שמתארים את העונש שהטיל השופט על הנאשם.
\item אל תכלול משפטים אשר מכילים ציטוטים לתיקים אחרים.
\end{enumerate}

\textbf{\# שלבים לביצוע Steps Reasoning}
\begin{enumerate}
\item קרא את פסק הדין המלא.
\item עבור כל משפט, שאל: האם יש כאן הסבר או נימוק שיפוטי או עקרון משפטי כלשהו? שנאמר ע"י השופט.
\item אם כן — סווג את המשפט לאחת הקטגוריות לעיל.
\item דלג על משפטים שלא מכילים נימוק או הסבר אמיתי.
\item החזר את כל המשפטים הרלוונטיים.
\end{enumerate}

\textbf{\# דוגמאות למשפטים שאינם תקפים:}
\begin{itemize}
\item ״מדובר באירוע חמור״ — זהו משפט טענתי, ללא נימוק.
\item ״יש להחמיר את הענישה״ — אין בו הסבר.
\item על פי ממצאי שירות המבחן הנאשם השתתף, במשך תקופה משמעותית, בהליך טיפולי מוצלח. — יש פה עובדה ללא הסבר.
\item אני מתיר לנאשם לרצות את עונש המאסר בן שלושה חודשים, מהם ינוכו 10 ימי מעצרו, בעבודות שירות.
\item המבקר קבע כי מתחם העונש ההולם נע בין 20 חודשי מאסר בפועל לבין 48 חודשי מאסר בפועל.
\end{itemize}

\textbf{\# דוגמאות למשפטים תקפים} (אין להחזיר אותם בפלט בפועל):
\begin{itemize}
\item ״עבירת הצתה הינה מן החמורות שבעבירות, שיודע אתה את תחילתה ואין אתה יודע את סופה.״
\item ״הנאשם ביצע את העבירה תוך הפרת אמון משמעותית שניתנה בו כמטפל.״
\item ״הפגיעה בנפגע העבירה חמורה ומתמשכת, ולכך יש לתת ביטוי בענישה.״
\item ״היתר הבנייה הקיים בהווה הנו נסיבה אשר יש להתחשב בה לקולא, שכן מדובר במבנה שהיום ברור לגביו כי הוא עומד בחוקי התכנון והבנייה הנוכחיים, ולפיכך, ברור כעת כי עמידתו במקומו אינה פוגעת בערכים חשובים שעניינם תכנון מוניציפלי, תכנון תוואי הנוף, הקצאת חלקות וכן זכויות קניין.״
\item ״אין ספק, כי הטלת ענישה ממשית עליו תפגע גם בו וגם בבני משפחתו.״
\item ״הנאשם הורשע לאחר שניהל משפטו ואמנם אין להעניש נאשם על עצם ניהול משפט פלילי, אך אין להתחשב בו כפי שמתחשבים באדם שהודה בפתח משפטו והביע חרטה כנה על מעשיו.״
\end{itemize}
}%
\end{quote}

\subsubsection*{Step 2: Judicial Reasoning Validation}
\textbf{Model:} GPT-4o-mini

\begin{quote}
\small
\foreignlanguage{hebrew}{%
השב \textbf{כן} אם המשפט שלפניך כולל הסבר או נימוק שמצדיק החלטה שיפוטית (למשל: הפניה לעקרון משפטי, שקלול נסיבות, נסיבת לקולא/לחומרה, מטרת הענישה).
השב \textbf{לא} אם המשפט הוא רק תיאור עובדות, הליך, ציטוט טענות או הצהרה תוצאתית ללא הסבר.

דוגמאות:

מטבע הדברים, לא ניתן לאותם מכתבים כל משקל. — לא

לטעמי במקרה זה נכון היה לחרוג ממתחם העונש ההולם, וזאת מטעמי צדק, גם במקרה שהמדינה לא פעלה במזיד או ברשלנות. — כן

הוא ונאשם 5 היו הקבלנים הפעילים והבולטים ביותר בפגישה. — לא

בעבירות הלבנת הון נפגעו הערכים המוגנים של שמירה על אמינות המערכת הפיננסית, כלכלת המדינה והשוויון. — כן

אני סבור שמתחם העונש ההולם בגין העבירה קרוב יותר למתחם שהציעה התביעה, קרי 8 חודשים ועד 18 חושדים. — לא

אל תוסף מילים; כתוב רק כן או לא. אם המשפט גבולי השב לא.
}%
\end{quote}

\bigskip
\subsubsection*{Step 3: Question Generation}
\textbf{Model:} GPT-4.1-mini

\begin{quote}
\small
\texttt{\#\#\# Role and Objective}

You are a seasoned Israeli jurist who writes precise, exam-style questions intended to surface the legal rationale found in short ``Answer'' sentences extracted from court verdicts.

\texttt{\#\#\# Instructions}

For every item you receive, generate \textbf{one} question in modern Hebrew that a law student could answer exactly with the given ``Answer.'' Keep each question:
\begin{itemize}
\item Focused on the key legal principle or factual nuance in the Answer.
\item Neutral in tone, neither accusatory nor apologetic.
\item One sentence, max 25 words.
\end{itemize}

Sub-categories:
\begin{enumerate}
\item \textbf{Language} --- Output questions in Hebrew.
\item \textbf{Style} --- Prefer \foreignlanguage{hebrew}{״כיצד״, ״מהו״, ״מדוע״} to open questions; avoid rhetoric.
\item \textbf{Scope} --- Use only the Answer (and optional Context to disambiguate names).
\item \textbf{Punctuation} --- Standard Hebrew punctuation; never the character `\texttt{-}'.
\end{enumerate}

\texttt{\#\#\# Reasoning Steps}
\begin{enumerate}
\item Identify its pivotal legal point (rule, fact pattern, mitigation, etc.).
\item Reformulate that pivot into an interrogative sentence that elicits the Answer verbatim or in tight paraphrase.
\item Verify the question is answerable solely from the Answer.
\end{enumerate}

\texttt{\#\#\# Output Format}

Return exactly one line per Answer, numbered \texttt{1.}~\texttt{2.}~\ldots

\textit{Example:}\\
\foreignlanguage{hebrew}{1. מהו השיקול המרכזי שהצדיק החמרה בענישת הנאשם?}\\

\foreignlanguage{hebrew}{2. כיצד השפיעה הודאת הנאשם בשלב מוקדם על מתחם העונש?}

Return only the numbered list---nothing else.
\end{quote}

\bigskip
\subsubsection*{Step 4: QA Validation}
\textbf{Model:} GPT-4o-mini

\begin{quote}
\small
\foreignlanguage{hebrew}{%
אתה בודק-איכות קפדן.
השאלה והתשובה חייבות להתאים בזיקה מלאה: האם התשובה עונה במדויק על השאלה ענה אך ורק ב-״1״ (כן, עונה) או ״0״ (לא).
}%
\end{quote}

\end{document}